\documentclass[journal]{IEEEtran}
\usepackage{cite} 
\usepackage{graphicx}
\usepackage{subfigure}
\usepackage{amsmath}
\usepackage{booktabs} 
\usepackage{amsfonts} 
\usepackage{nicefrac} 
\usepackage{microtype} 

\usepackage[noend]{algpseudocode}
\usepackage{multirow}
\usepackage{algorithmicx,algorithm}
\ifCLASSINFOpdf
\else
\fi
\begin{document}

%
\title{ Camera-Invariant Meta-Learning Network for Single-Camera-Training Person Re-identification}
%
%
%

\author{Jiangbo~Pei\IEEEauthorrefmark{1},
Zhuqing~Jiang\IEEEauthorrefmark{2}\IEEEauthorrefmark{1},
Aidong~Men,
Haiying~Wang,
Haiyong~Luo,
Shiping~Wen
\thanks{Jiangbo Pei, Aidong Men and Haiying Wang are with the School of Artificial Intelligence, Beijing University of Posts and Telecommunications, Beijing 100876, China, (e-mail: jiangbop@bupt.edu.cn; menad@bupt.edu.cn; why@bupt.edu.cn).}
\thanks{Zhuqing Jiang is with Beijing Key Laboratory of Network System and Network Culture, and also with the School of Artificial Intelligence, Beijing University of
Posts and Telecommunications, Beijing 100876, China, (e-mail: jiangzhuqing@bupt.edu.cn).}%
\thanks{Haiyong Luo is with Institute of Computing Technology, Chinese Academy of Sciences, Beijing 100190, China, (e-mail: yhluo@ict.ac.cn).}%
\thanks{Shiping Wen is with Australian AI Institute, Faculty of Engineering and Information Technology, University of Technology Sydney, NSW 2007, Australia, (email: shiping.wen@uts.edu.au).}%
\thanks{\IEEEauthorrefmark{2}Corresponding author}%
\thanks{\IEEEauthorrefmark{1}Equal contribution}}

	\maketitle

\begin{abstract}
Single-camera-training person re-identification (SCT re-ID) aims to train a re-ID model using SCT datasets where each person appears in only one camera. The main challenge of SCT re-ID is to learn camera-invariant feature representations without cross-camera same-person (CCSP) data as supervision. Previous methods address it by assuming that the most similar person should be found in another camera. However, this assumption is not guaranteed to be correct. In this paper, we propose a Camera-Invariant Meta-Learning Network (CIMN) for SCT re-ID. CIMN assumes that the camera-invariant feature representations should be robust to camera changes. To this end, we split the training data into meta-train set and meta-test set based on camera IDs and perform a cross-camera simulation via meta-learning strategy, aiming to enforce the representations learned from the meta-train set to be robust to the meta-test set. With the cross-camera simulation, CIMN can learn camera-invariant and identity-discriminative representations even there are no CCSP data. However, this simulation also causes the separation of the meta-train set and the meta-test set, which ignores some beneficial relations between them. Thus, we introduce three losses: meta triplet loss, meta classification loss, and meta camera alignment loss, to leverage the ignored relations. The experiment results demonstrate that our method  achieves comparable performance with and without CCSP data, and outperforms the state-of-the-art methods on SCT re-ID benchmarks. In addition, it is also effective in improving the domain generalization ability of the model.
\end{abstract}

\IEEEpeerreviewmaketitle
\section{Introduction}
\IEEEPARstart{P}{erson} re-identification (re-ID) aims to identify a specific person across multiple camera views \cite{bai2021person30k,li2014clothing,zheng2018subspace,wang2014camera,ning2020feature,si2017spatial,li2020hierarchical,bai2021sanet,li2019attribute,hu2022divide,nodehi2021multi,li2022cocas+,tan2021incomplete,mclaughlin2016person,zhang2021unrealperson,zhao2021learning,bai2021unsupervised,li2021combined,chen2021learning,yang2021joint}.
It has drawn increasing attention in recent years due to its importance in video surveillance and security systems.
Most previous works focus on supervised re-ID \cite{zhu2020identity,zhang2020relation,chen2020salience,an2015person,tan2016dense,li2016person,he2021partial,zhang2021coarse}. Although they have achieved high accuracies, their success relies on tremendous annotated data. More specifically, it depends particularly on the annotated data that belongs to the same person but is captured from different cameras, which we call cross-camera same-person (CCSP) data.
\begin{figure}
\centering
\includegraphics[width=0.9\linewidth]{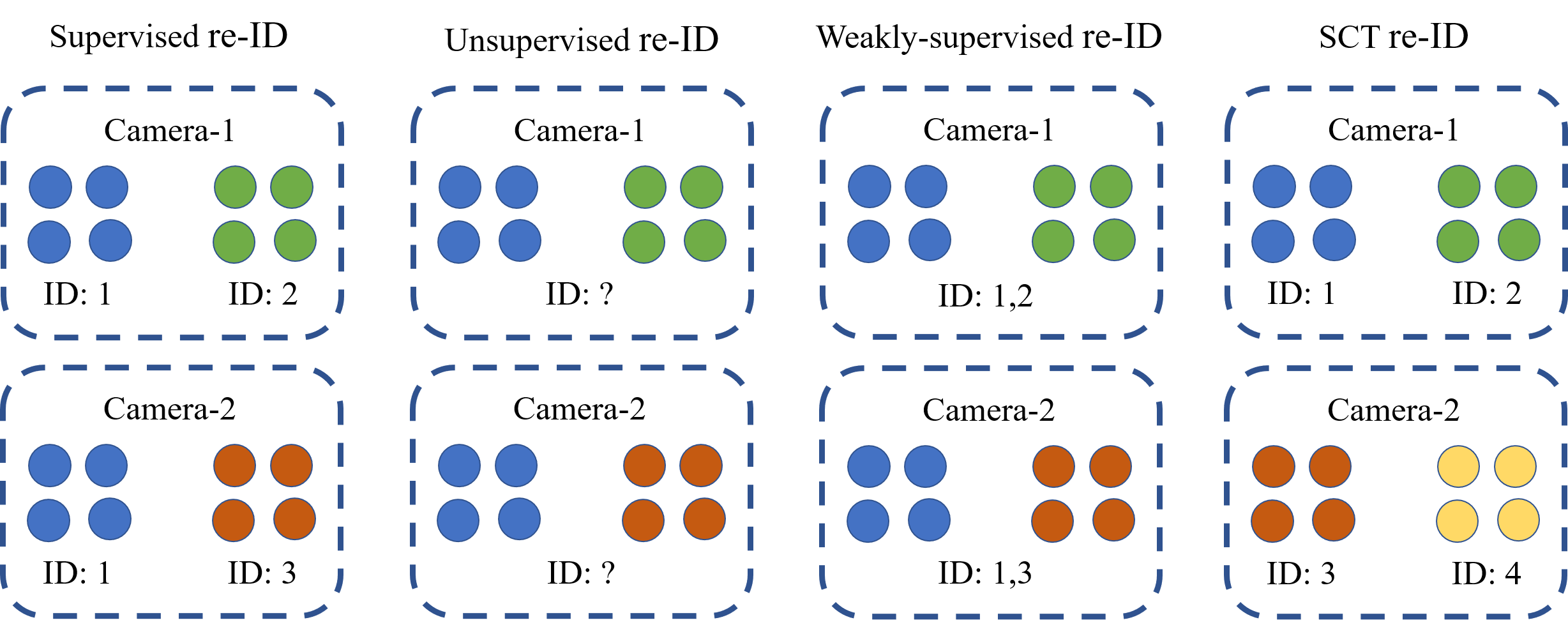}%
\vspace{-2mm}
\caption{The comparison between SCT re-ID setting and previous re-ID settings. Different colors represent different identities.
Conventional supervised re-ID data are composed of a large number of people appearing in multiple cameras, and there are many data that belong to the same identity but are captured by different cameras (CCSP data), such as the ID-1 in this figure. In this setting, we know the identity of each data.
Unsupervised re-ID has no identity annotation for data, but it still requires latent CCSP data in the training set (such as the blue data).
In weakly-supervised re-ID, data are usually weakly annotated. We showed an example of \cite{wang2019weakly}, which annotates data at bag-level, and the identity of each data is unknown. Similar to unsupervised re-ID, this setting also needs CCSP data for training.
In contrast, under the single-camera-training (SCT) setting, each person appears in only one camera; thus, there are no CCSP data. Compared to previous settings, this setting is easy to collect and annotate data.
}\label{setting}
\vspace{-4mm}
\end{figure}
In practice, it is tough to collect so much available CCSP data. First, the collection is based on the assumption that there is a large amount of overlapping person between different cameras. However, this assumption may not be valid in the real world, especially between some remote cameras. Second, since the time of a person appearing in different cameras is unpredictable, searching and annotating CCSP data are exhausting and expensive projects.
These limitations severely hinder the real-world, large-scale deployment of conventional supervised re-ID methods.

In the past few years, many works have been proposed to reduce the annotation cost. These methods belong to three main categories: namely unsupervised re-ID, semi-supervised re-ID, and weakly-supervised re-ID. Unsupervised re-ID methods \cite{fu2021unsupervised,lin2020unsupervised,wang2020camera} aim to train a re-ID model using only unlabeled data, while semi-supervised re-ID methods \cite{xin2019semi} add some labeled data as assistance. And for weakly-supervised re-ID \cite{wang2019weakly, chen2021joint,zhu2019intra}, previous works focus on leveraging more economic annotation as supervision. Although these methods have achieved outstanding results in alleviating the annotation cost, they do not change the inherent reliance of the model on CCSP data.
Thus, they still require a lot of unlabeled/weakly-labeled CCSP data for training.

In this paper, we aim to address the single-camera-training (SCT) re-ID \cite{zhang2020single}, which eliminates the reliance of the re-ID model on CCSP data by enforcing each person appears in only one camera. We compare SCT re-ID setting to previous settings in Fig. \ref{setting} and illustrate their two advantages as follows:
\textbf{First}, SCT re-ID removes the heavy burden of collecting CCSP data and makes it easy to prepare the training data. For example, using some existing tracking techniques \cite{keuper2018motion,luo2018trajectories}, we can quickly get a large number of tracklets under each camera at different periods, and each of them most likely corresponds to a unique ID.
\textbf{Second}, SCT re-ID meets the fact that there is almost no overlapping identity between remote cameras. Thus, it has the potential to be deployed in more scenarios.


The main challenge of SCT re-ID is to learn camera-invariant feature representations without CCSP data. Specifically, previous re-ID settings have sufficient CCSP data for training, which provides the critical supervision for a model to learn camera-invariant feature representations. In contrast, there are no available CCSP data in SCT re-ID, so that we must turn to other types of supervision to achieve the goal.
The prior work \cite{zhang2020single} assumes that the most similar person is found in another camera. Based on the assumption, they formulate a multi-camera negative loss function to build a relation between samples from different cameras. Although they have made progress, the irrationality of the assumption limits their performance in practice. 

To this end, we propose a Camera-Invariant Meta-Learning Network (CIMN) for SCT re-ID.
Intuitively, if the feature representations are robust to camera changes, we could consider them to be camera-invariant. Inspired by this, a cross-camera simulation via meta-learning pipeline are designed. Specifically, at each training iteration, CIMN random samples data from two different cameras as meta-train set and meta-test set, respectively.
CIMN first virtually trains the meta-train set to learn feature representations and then transfers the learned representations to the meta-test set to examine their applicability on the new set.
The objective of the cross-camera simulation named simulation loss, enforcing the representations learned from the meta-train set to perform well on the meta-test set.
With the cross-camera simulation, CIMN can learn camera-invariant and identity-discriminative representations even there are no CCSP data.
However, this simulation also causes the separation of the meta-train set and the meta-test set, which ignores some beneficial relations between them. A natural relation between the two sets is the negative relations, i.e., the images in one set have a different identity from the images in the other set since they belong to different cameras. To leverage this relation, we design two loss functions, namely meta triplet loss, and meta classification loss.
There are also positive relations between the two sets. Specifically, we argue that there should be no camera-related features in the camera-invariance feature representations. Thus, data from different sets should be embedded into the same feature space. To exploit this relation, we introduce a meta camera alignment loss to align the distribution of the two sets in the feature space. Finally, these losses and the simulation loss are combined for meta-optimization.
Note that the meta-learning technology has been introduced into re-ID currently: \cite{choi2021meta} propose MetaBIN to simulate unsuccessful generalization scenarios for domain generalizable re-ID, and \cite{yang2021joint} exploit meta-learning pipeline for unsupervised re-ID. However, our purpose is entirely different from them because they have a lot of labeled/unlabeled CCSP data for training, while we aim to train the model without CCSP data.

We conduct extensive experiments to demonstrate the effectiveness of our proposed approach on three open-source SCT re-ID benchmarks.
The result shows that our approach is superior to the state-of-the-art SCT re-ID methods. More valuable, our controlled experiment shows that \emph{ under the same data volume, the performance of our CIMN  without CCSP data is comparable to that of the conventional approaches with CCSP data, and is also similar to that of CIMN with CCSP data}. In addition, by promoting to learn camera-invariant feature representations, CIMN is also robust to camera changes. As a result, our model is also more generalizable to new re-ID datasets captured from other cameras.

We summarize our contributions as follows:
\begin{itemize}
\item In this paper, we propose a Camera-Invariant Meta-Learning Network (CIMN) to address the challenging SCT re-ID task. CIMN proposes a cross-camera simulation process to enforce the representations learned from one camera (the meta-train set) to perform well on another camera (the meta-test set), thus guiding the model to learn camera-invariant representations in the absence of cross-camera same-person  (CCSP) data.

\item  We explore the negative and positive relations that are ignored in the cross-camera simulation process and propose the meta triplet loss, the meta classification loss and the meta camera alignment loss to exploit these relations.

\item Extensive experiments demonstrate the effectiveness of our method in SCT re-ID. Particularly, we show that the performance of our CIMN  without CCSP data is comparable to that of the conventional approaches with CCSP data under the same data volume. Besides, our method also promotes the model's generalization ability to new re-ID datasets collected from other cameras.

\end{itemize}

\begin{figure*}
\centering
\includegraphics[width=0.9\linewidth]{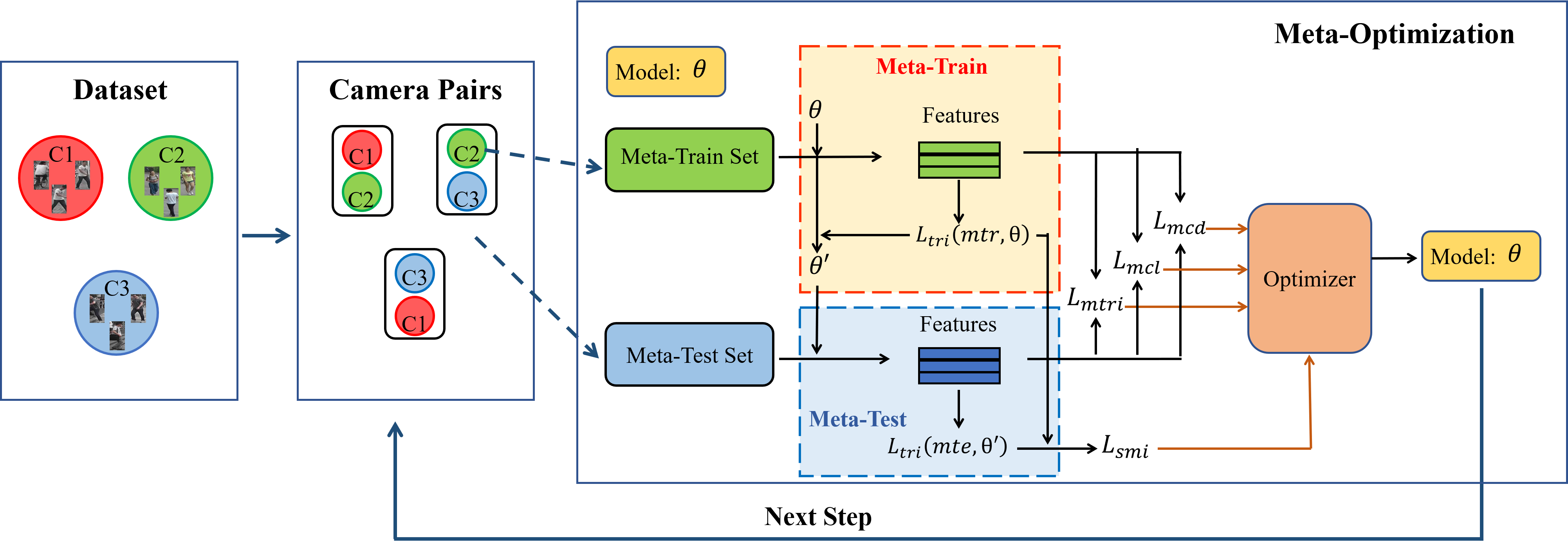}%
\caption{An overview of our proposed CIMN. Three cameras are presented in this figure for a demonstration. Before training, we first construct camera pairs. Then we build meta-train set and meta-test set from each pair to construct meta-batch. During training, CIMN cuts off the relation between two sets to perform cross-camera simulation with the meta-train process and the meta-test process, calculating the simulation loss $L_{smi}$. After that, the meta triplet loss $L_{mtri}$, the meta classification loss $L_{cl}$, and the meta camera alignment loss $L_{mca}$ are imposed to leverage the negative and positive relations between the two sets. Finally, all losses are combined to optimize the model.
\vspace{-4mm}
}\label{net}
\end{figure*}
\vspace{-3mm}
\section{Related Work}
\label{sec:2}
\subsection{Conventional Person Re-Identification}

Conventional supervised re-ID is often referred to as fully-supervised re-ID \cite{liu2021learning,zhang2020single,wang2022nformer,mclaughlin2017video,ren2021learning,gu2022autoloss,wang2021robust,liu2019bayesian,chai2022video,subramanyam2023meta,jin2021occlusion,shu2021large,zhang2021person}. Most of the conventional supervised re-ID works aim to learn identity-discriminative features with representation methods \cite{wu20193}. These methods design attention selection\cite{chen2019abd}, semantic segmentation\cite{zhu2020identity} and diverse representations \cite{zhang2020relation,chen2020salience} for extracting robust features.
Besides, many works \cite{rao2020improved}  address conventional supervised re-ID using metric learning methods. Inspired by  advanced metric learning works \cite{musgrave2020metric}, lots of ranking algorithms\cite{hao2019hsme} and metric loss functions \cite{zhao2020deep,sun2020circle} are proposed for re-ID.
Although existing re-ID studies have achieved impressive accuracies in the field of the conventional supervised re-ID, their successes heavily rely on the annotated CCSP data. As mentioned, CCSP data is expensive to be annotated and is few to none among remote cameras, which prevents these methods from deploying on a large scale.

In the past few years, many works are proposed to reduce the annotation costs of re-ID. Unsupervised re-ID \cite{han2022delving,xuan2021intra,wang2021camera,qi2021adversarial,lin2020unsupervised,wang2020camera,dai2021dual,li2020joint,wang2021central,wang2022cloning,liao2022graph,cho2022part} aims to train a re-ID model only using unlabeled data. These methods usually assign pseudo labels to unlabeled data and then train the model with the pseudo labels. Semi-supervised re-ID \cite{xin2019semi} uses labeled and unlabeled data to learn a re-ID model jointly. Similar to the unsupervised methods, they also focus on pseudo-labels approaches, but with additional labeled data as an auxiliary.
Weakly-supervised re-ID aims to use more economic annotation to replace the full annotation in conventional supervised re-ID.
For example, Wang \textit{et al.} \cite{wang2019weakly} group the images into bags and annotate data with bag-level annotations.
Zhu \textit{et al.} \cite{zhu2019intra} propose to annotate each camera independently.
Although the above methods have achieved great results in reducing the annotation cost, their essence is to self-discover the identity correspondence among unlabeled/labeled+unlabeled/weakly-labeled data, which still requires potential CCSP data for training.

\subsection{Single-Camera-Training Person Re-Identification}
SCT re-ID\cite{zhang2020single} assumes that there is no CCSP data in the training set.
It not only helps to reduce the annotation costs, but also has the potential to be deployed to some remote cameras where there is no CCSP data.

To address SCT re-ID, Zhang \textit{et al.} \cite{zhang2020single} assume that the most similar person is found in another camera to associate data from different cameras. However, since the changes in the camera often lead to larger discrepancies in the image, this assumption is hardly guaranteed to hold true.
Ge \textit{et al.} \cite{ge2021cross} design a local-branch that localize and extract local-level feature for SCT re-ID using a transformer \cite{vaswani2017attention} structure.
Wu \textit{et al.} \cite{wu2022camera} propose CCFSG which introduce the $\sigma-$Regularized Conditional Variational Autoencoder to  synthesize the cross-camera samples for model training. 
In contrast, we propose the CIMN that guides the feature representation to be camera-invariant by enforcing the representation learned from a particular camera to perform well on other cameras, which is model-agnostic and is orthometric to these methods \cite{ge2021cross, wu2022camera}.


\subsection{Meta-Learning }
\label{sec:meta-learning}
Meta-Learning has recently obtained much attention, in which a learner learns new tasks and another meta learner learns to train the learner. Recent meta-learning studies can be roughly classified into optimizing-based\cite{finn2017model}, model-based\cite{santoro2016meta}, and metric-based methods \cite{sung2018learning}.
Our work is related to Model Agnostic Meta-Learning (MAML) \cite{finn2017model}, an optimizing-based meta-learning framework that aims to find a good initialization of parameters that are sensitive to changes in novel tasks. The variants of MAML have been widely used in computer vision tasks various tasks, such as domain generalization \cite{li2018learning}, and frame interpolation \cite{choi2020scene}. Recently, MAML has also been introduced into re-ID, including domain generalized re-ID \cite{choi2021meta,ni2022meta} and unsupervised re-ID\cite{yang2021joint}. Choi \textit{et al.} \cite{choi2021meta} and Ni \textit{et al.} \cite{ni2022meta} aim at simulating domain generalization scenarios to generalize normalization layers, which focuses on the domain level discrepancy. In both their meta-train process and meta-test process, the model is trained on several fully labeled datasets with sufficient CCSP data.
Yang \textit{et al.} \cite{yang2021joint} are more related to our works, which also focus on the camera level. They aim to adapt to the shifts caused by cameras by simulating a cross-camera searching process.
However, the searching process essentially relies on associating potential unlabeled CCSP data with the cross-entropy loss function based on the pseudo-labels.
In contrast, we mainly focus on how to train the model without CCSP data. We deliberately cut  the relations between different cameras during the cross-camera simulation, and supplement them with  three proposed losses in subsequent meta-optimizations. Thereby, our method is different from \cite{choi2021meta,yang2021joint} in both purpose and implementation.

\subsection{Domain Generalizable Person Re-Identification }
Current re-ID models trained on a source dataset (domain) are well known to suffer from performance degradation when generalized to a new target dataset.
To this end,  a lot of generalizable re-ID methods are proposed \cite{choi2021meta,ni2022meta,song2019generalizable,jin2020style,chen2020dual,liu2022debiased,liao2021transmatcher,han2022generalizable,dai2021generalizable}. For example, Jin \textit{et al.}\cite{jin2020style} design an effective SNR module to distill identity-relevant feature from the removed information and restitute it to the network. Chen \textit{et al.} \cite{chen2020dual} propose a DDAN to map images into a domain-invariant feature space by selectively aligning the distributions of multiple source domains. Although domain generalization ability is not our main focus, we find that our model is also beneficial to this field. We conjecture the reason is that CIMN encourages the model to be robust to camera changes, therefore promoting it to be more generalizable to new re-ID datasets captured from other cameras.

\vspace{-3mm}
\section{APPROACH}
\label{sec:3}
\vspace{-2mm}
\subsection{Overview}

We begin with a formal description of the SCT re-ID problem. We assume that we are given a dataset D captured from $N$ cameras $D=\{{C_n}\}_{n=1}^N$. Each camera $C_n$ contains $M_n$ image-label pairs $C_n=\{(x_i^n,y_i^n)\}_{i=1}^{M_n}$. Each image $x_i^n\in \mathcal{X}_n$ corresponds to an identity label $y_i^n \in \mathcal{Y}_n$. Different from conventional re-ID, there is no overlapped identity between two different cameras in SCT re-ID, i.e., $\forall i,j, 0\le i<j\le N : \mathcal{Y}_i \cap \mathcal{Y}_j = \emptyset$. Our goal is to learn a model parameterized by $\theta$ with $D$,
which can extract camera-invariant identity-discriminative feature $f\in \mathbb{R}^{d \times 1}$ for cross-camera retrieval. 
The overall of our proposed CIMN is presented in Fig. \ref{net}.
Particularly,  we  construct camera pairs and sample meta-train set and meta-test set from each pair to form meta-batch. 
During training, CIMN  cuts off the relation between two sets to perform a cross-camera simulation, and encourages the model to learn camera-invariant features by the simulation loss $L_{smi}$. After that, the meta triplet loss $L_{mtri}$, the meta classification loss $L_{cl}$, and the meta camera alignment loss $L_{mca}$  are proposed to leverage the negative and positive relations that ignored by the cross-camera simulation.

\vspace{-3mm}
\subsection{Meta-Batch Preparation.}
To simulate a cross-camera process, we sample meta-batch that includes images from two different cameras, and assign them to meta-train set and meta-test set according their camera IDs.
Given a training dataset captured by $N$ different cameras $D=\{C_1, C_2, ..., C_N \}$, we sample the meta-batch as follows: (1) In the $n$-th epoch, we select $C_{q(n)}$ as the meta-train camera $C_{mtr}$, where $q(n)$ is the remainder of $n$ divided by $N$. (2) A meta-test camera $C_{mte}$ is randomly selected from the remaining cameras, forming camera pair $\{C_{mtr},C_{mte}\}$. (3) We randomly sample $P\times K$ images from the meta-train camera as the meta-train set $S_{mtr}$ and $P\times K$ images from the meta-test camera as the meta-test set $S_{mte}$, where $P$ is the number of identities, and $K$ is the number of images that belong to each identity. Then, a $2\times P\times K$ meta-batch is built, denoted as $\{S_{mtr},S_{mte}\}$. 
\vspace{-3mm}
\subsection{Cross-Camera Simulation}
\label{kccall}
The cross-camera simulation simulates a process that trains a model on one camera and evaluates its performance on a new camera via meta-learning.
The process consists of two steps: meta-train and meta-test.
\subsubsection{Meta-Train}
In each meta-batch $\{S_{mtr},S_{mte}\}$, given a model with initial parameters $\theta$, we first virtually train the model on the meta-train set $S_{mtr}$.
In this step, we adopt triplet loss\cite{hermans2017defense} as criterion. 
The triplet loss on $S_{mtr}$ can be formulated as:
\begin{equation}
\label{mtr_tri}
L_{tri}(mtr,\theta)=\sum_{j=1}^{P\times K}[ d(f_j,f_{j}^p))-d(f_j,f_{j}^n))+m]^+,
\end{equation}
where $f_j=f_{\theta}(x_j)$ represents the feature of an anchor image $x_j$ extracted by the model with parameter $\theta$, $f_j^p$ and $f_j^n$ are the features of the anchor's farthest positive image and nearest negative image within $S_{mtr}$, $d(.)$ is the euclidean distance function, $m$ is a margin value set to 0.3, and $[z]^+=max(z,0)$.

With a learning rate $\eta$, we can update the model with SGD optimizer like the conventional metric learning. Then we obtain an intermediate parameter $\theta '$, representing the parameter of the model after learning on the $S_{mtr}$. 
\begin{equation}
\label{update}
\theta '=\theta-\eta \nabla_{\theta} L_{tri}(mtr,\theta).
\end{equation}

\subsubsection{Meta-Test}
We then transfer the  model trained on $S_{mtr}$ (parameterized by $\theta '$) to the meta-test set $S_{mte}$ for evaluation.
Following \ref{mtr_tri}, we calculate the triplet loss of the model on $S_{mte}$:
\begin{equation}
\label{mte_tri}
\begin{aligned}
L_{tri}(mte,\theta')= \sum_{j=1}^{P\times K}[ d({f'}_j,{f'}_{j}^p))-d({f'}_j,{f'}_{j}^n))+m]^+,
\end{aligned}
\end{equation}
where ${f'}_j=f_{\theta'}(x_j)$ represent the feature of the image $x_j$ extract by the model with the intermediate parameter $\theta'$.

\subsubsection{Simulation Loss}
On the one hand, we expect the model to learn identity-discriminative representations from $S_{mtr}$ so that it should perform well on this set. On the other hand, we expect the model learned from $S_{mtr}$ to be camera-invariant, so that it should also perform well on $S_{mte}$.
With the two purposes,  the objective of the cross-camera simulation can be formulated as:
\begin{equation}
\label{simobject}
L_{smi}=\lambda L_{tri}(mtr,\theta) + (1-\lambda) L_{tri}(mte,\theta'),
\end{equation}
where $\lambda$ is a trade-off coefficient of the two purposes.

By substitute Eq. \ref{update} into this objective, the simulation loss is:
\begin{equation}
\label{simobject1}
\begin{aligned}
L_{smi}=&\lambda L_{tri}(mtr,\theta) \\+& (1-\lambda) L_{tri}(mte, \theta-\eta \nabla_{\theta} L_{tri}(mtr,\theta)).
\end{aligned}
\end{equation}

Here we explain why we adopt triplet loss in the cross-camera simulation. Although there is no overlapping person between the meta-train set and the meta-test set, the two sets still have potential relations. We should cut off these relations in meta-train to avoid the model implicitly touching the meta-test set. We select the triplet loss as our criterion considering its local-level nature, i.e., it only considers the relationship between samples in a certain closed set and does not interact with other sets.

On the other hand, although we have to cut off the relations between the meta-train set and the meta-test set during simulation, these relations are undoubtedly beneficial for training. To leverage these relations, we further design three losses: 1) meta triplet loss, 2) meta classification loss and 3) meta-camera alignment loss, which are introduced in the following section.
\vspace{-3mm}
\subsection{Meta-Optimization}

\subsubsection{Meta Triplet Loss}
We introduce a meta triplet loss to capture the identity relations between the two sets. In SCT re-ID setting, these relations are the negative relations, i.e., the images in one set have completely different identities from the images in the other set.
Specifically, for an image, meta triplet loss aims to pull it closer to its farthest positive sample in the same set, and push it away from the nearest negative sample in different sets. It can be formulated as:
\begin{equation}
\label{mtri}
\begin{aligned}
L_{mtri}= \sum_{j=1}^{2\times P\times K}[ d({F}_j,{F}_{j}^p))-d({F}_j,{F}_{j}^n))+m]^+,
\end{aligned}
\end{equation}
where $F_j$ is the feature of an anchor image $x_j$, $F_j^p$ is the feature of the anchor's farthest positive image within the same set, while $F_j^n$ is the feature of its nearest negative image in the other set. To be consistent with the cross-camera simulation, we use different parameters to extract features from the meta-train set and the meta-test set:
\begin{equation}
\label{F}
F_j=\left\{
\begin{aligned}
&f_{\theta}(x_j),& x_j\in S_{mtr}\\
&{f}_{\theta'}(x_j),& x_j\in S_{mte}\\
\end{aligned}
\right.
\end{equation}

\subsubsection{Meta Classification Loss}
Classification loss is prevalent in the re-ID field, which uses cross-entropy to measure the difference between model prediction distribution and the actual label distribution. This loss has been widely demonstrated to be effective in re-ID.
It can naturally exploit the negative relations between the meta-train set and the meta-test set in the label space since an image that belongs to one identity label must not belong to other identity labels. It can not be adopted by the cross-camera simulation which requires cutting off the two sets. However, this loss can be a good choice for associating the two sets.
To make it suitable for our model, we improve it to meta classification loss.

For the meta-train set $S_{mtr}$, we calculate the classification loss with the initial parameter $\theta$ in the meta-train process:
\begin{equation}
L_{cl}({mtr,\theta})=\sum_{j=1}^{P\times K}CE[(f_{\theta}(x_j)),y_j],
\end{equation}
where CE is the cross-entropy function, $x_j$ is the images in $S_{mtr}$, $y_j$ is the label of $x_j$, $f_{\theta}(x_j)$ represents the features of $x_j$ extracted by model $\theta$. Note that the features in this formulation should have the same dimension as the number of labels (identities).

For the meta-test set, the classification loss is calculated based on the intermediate parameters $\theta'$:
\begin{equation}
L_{cl}({mte,\theta'})=\sum_{j=1}^{ P\times K}CE(f_{\theta'}(x_j)),y_j],
\end{equation}
where $x_j$ is the images in $S_{mte}$.

Then the loss can be formulated as:
\begin{equation}\
\label{mcl}
L_{mcl}=L_{cl}({mtr,\theta})+L_{cl}({mte,\theta'}),
\end{equation}

\subsubsection{Meta Camera Alignment Loss}
\label{dcc}

Besides the negative relations in identities, there are also positive relations between the two sets. Specifically, we argue that there should be no camera-related features in the camera-invariance feature representations. Thus, data from different sets should be embedded into the same feature space. To this end, we propose a meta camera alignment loss, aiming at aligning the feature distribution of the two sets.
Meta camera alignment adopts Maximum Mean Discrepancy (MMD) \cite{gretton2012kernel} and center distance to measure the distance between two distributions.
We use the initial parameter $\theta$ as the model parameter to extract features from $S_{mtr}$ and use the intermediate parameter $\theta'$ to extract from $S_{mte}$. We denote the  feature sets  of $S_{mtr}$ and $S_{mte}$ as $F_{mtr}$ and $F_{mte}$, respectively.
The MMD is calculated by
\begin{equation}
\begin{aligned}
MMD(&F_{mtr},F_{mte})= \frac{1}{(P\times K)^2}[\sum_{i=1}^{P\times K}\sum_{j=1}^{P\times K}k(f_{mtr}^i,f_{mtr}^j)+\\& \sum_{i=1}^{P\times K}\sum_{j=1}^{P\times K}k(f_{mte}^i,f_{mte}^j)-\sum_{i=1}^{P\times K}\sum_{j=1}^{P\times K}k(f_{mtr}^i,f_{mte}^j)],
\end{aligned}
\end{equation}
Where and $k(.,.)$ is a Gaussian kernel, $f_{mtr}$, $f_{mte}$ is the features from $F_{mtr}$ and $F_{mte}$, respectively.

The center distance is the Euclidean distance between the center of $F_{mtr}$ and $F_{mte}$:
\begin{equation}
\begin{split}
Cen(F_{mtr},F_{mte})=\frac{1}{(P\times K)^2}\|\sum_{i=1}^{P\times K}f_{mtr}^i-\sum_{j=1}^{P\times K}f_{mte}^j\|^2,
\end{split}
\end{equation}

Then the meta camera alignment loss of two distribuitons can be formulated as
\begin{equation}
\label{mca}
\begin{aligned}
L_{mca}=MMD(F_{mtr},F_{mte})+Cen(F_{mtr},F_{mte}).
\end{aligned}
\end{equation}

\subsubsection{Overall}
Combined the three losses and simulation loss, the final objective of the meta-optimization can be formulated as:
\begin{equation}
\label{final}
\begin{aligned}
L=L_{smi}+\gamma_1 L_{mtri}+\gamma_2 L_{mcl}+ \gamma_3 L_{mca},
\end{aligned}
\end{equation}
where $\gamma_1$, $\gamma_2$, $\gamma_3$ represent the weight of meta triplet loss, meta classification loss and meta camera alignment loss. The whole process is illustrated in  Algorithm.
\ref{algorithm}.


\begin{algorithm}[t]
\caption{CIMN algorithm}
\label{algorithm}
\hspace*{0.02in} {\bf Input:}
Training dataset captured by $N$ cameras \\
\hspace*{0.45in} $D=\{C_1,C_2,...,C_N\}$. \\
\hspace*{0.02in} {\bf Init:}
Model $\theta$, hyperparameters $P$, $K$, $\eta$, $\lambda$, $\gamma_1$, $\gamma_2$, $\gamma_3$.
\begin{algorithmic}[1]
\For{$epoch$ in $max\_epoch/N$}
\For{each $C_{mtr}$ in $D$:}
\State Sampling $C_{mte}$ from remain cameras
\State Build camera pair $\{C_{mtr},C_{mte}\}$
\State{Preparing meta-batch }
\For{each meta-batch $\{S_{mtr},S_{mte}\}$}
\State{\bf Meta-Train:}
\State Calculate meta-train loss $L_{tri}(mtr,\theta)$
\State Update model $\theta '=\theta-\eta \nabla_{\theta} L_{tri}(mtr,\theta)$
\State{\bf Meta-Test:}
\State Calculating meta-test loss $L_{tri}(mte,\theta')$.
\State Calculating simulation loss:
\State $L_{smi}=\lambda L_{tri}(mtr,\theta) + (1-\lambda) L_{tri}(mte,\theta')$.
\State Calculating meta triplet loss $L_{mtri}$.
\State Calculating meta classificaiton loss $L_{mcl}$.
\State Calculating meta camera alignment loss $L_{mca}$.
\State{\bf Optimization:}
\State $L=L_{smi}+\gamma_1 L_{mtri}+\gamma_2 L_{mcl}+ \gamma_3 L_{mca}$.
\State Update $\theta\leftarrow \theta- \eta\frac{\partial L}{\partial \theta}$.
\EndFor\State endfor
\EndFor\State endfor
\EndFor\State endfor

\end{algorithmic}
\end{algorithm}

\begin{figure}
\centering
\includegraphics[width=0.99\linewidth]{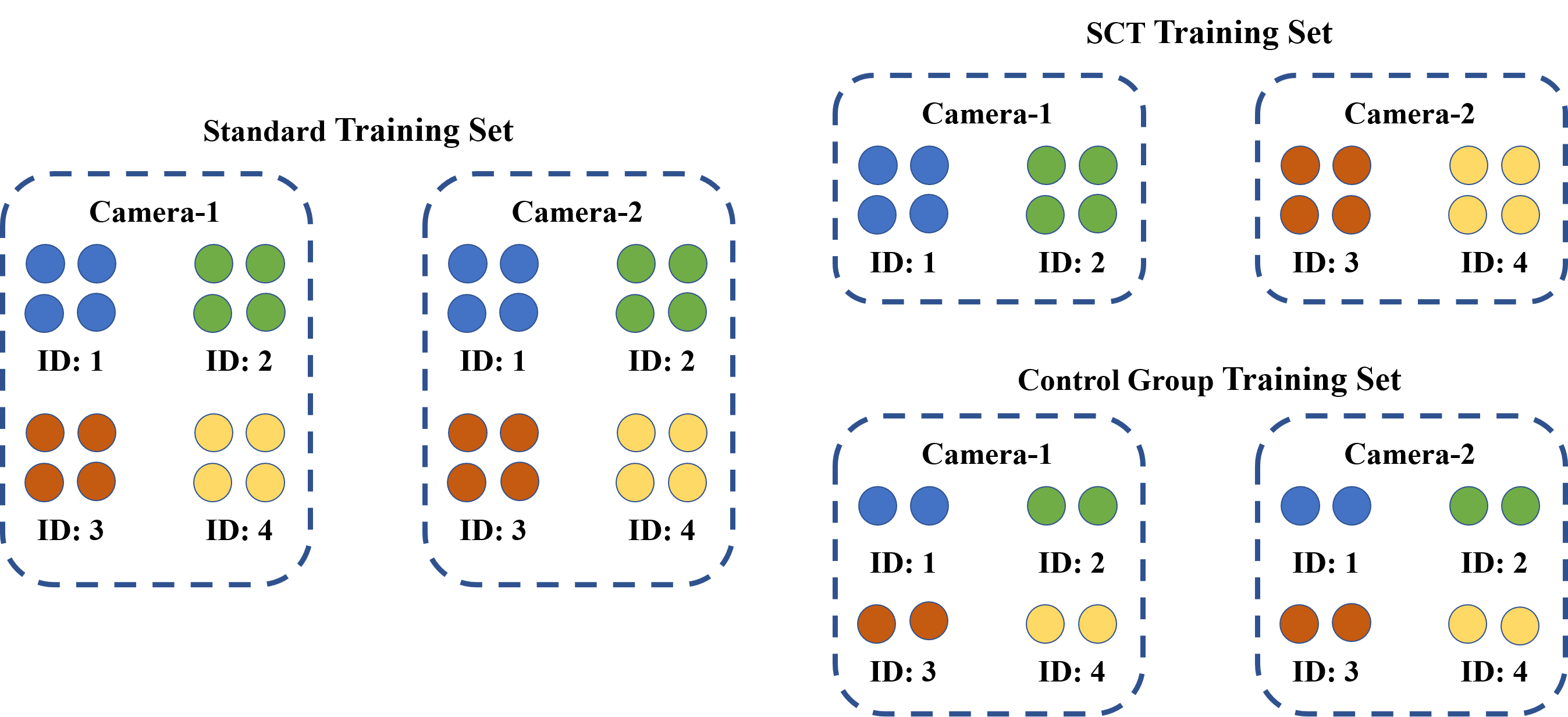}%
\vspace{-2mm}
\caption{The constructions of the training set of the standard setting (Market-STD and Duke-STD), SCT setting (Market-SCT and Duke-SCT) and the control group setting (Market-CG and Duke-CG).
\vspace{-2mm} 
}\label{sct_illu}
\end{figure}

\begin{figure}
\centering
\includegraphics[width=0.8\linewidth]{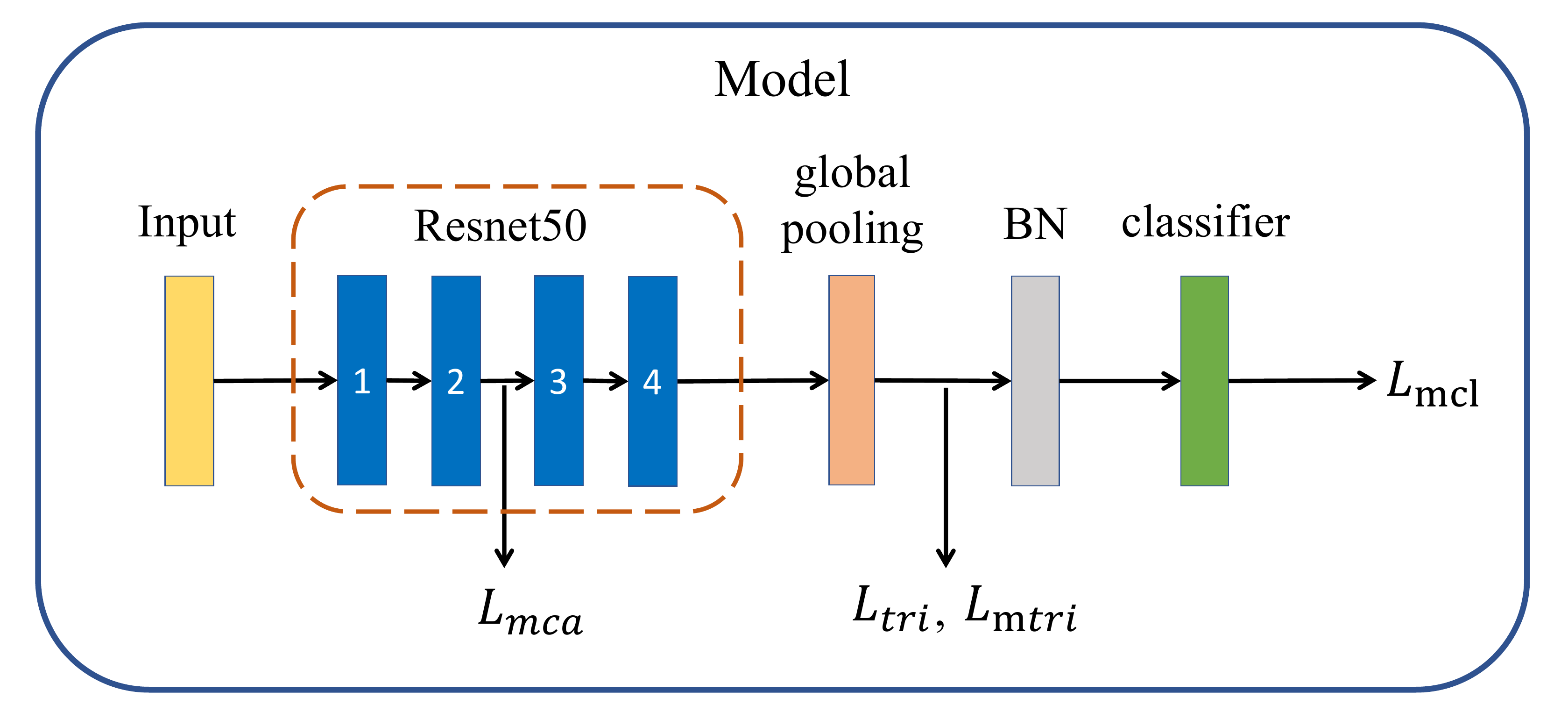}%
\vspace{-2mm}
\caption{The structure of our baseline model. 
\vspace{-4mm}
}\label{modelst}
\vspace{-3mm}
\end{figure}

\begin{table}
\caption{ Detailed statistics of the datasets used in our experiments. Market-STD, Market-SCT and Market-CG represent Market-1501 dataset in standard protocol, SCT protocol and Control Group protocol, respectively. The same is true for Duke-STD, Duke-SCT, Duke-CG, MSMT-STD and MSMT-CG. }
\label{dataset}
\centering
\resizebox{\linewidth}{!}{
\begin{tabular}{lccccc}
\toprule
Dataset &Train IDs&Train Images&Test IDs&Test Images & With CCSP data?\\

\midrule

Market-STD & 751&12,936&750&15,913&True \\
Market-SCT &751&3,561&750&15,913&False\\
Market-CG&751&3,561&750&15,913&True\\
Duke-STD & 702& 16,522&1,110&17,661&True\\
Duke-SCT & 702& 5,993&1,110&17,661&False\\
Duke-CG & 702& 5,993&1,110&17,661&True\\
MSMT-SCT & 1041& 6.645&3,060&93.820&False\\
\bottomrule
\end{tabular}}
\end{table}

\vspace{-3mm}
\section{Experiments}
\label{sec:5}

\subsection{Experiment Settings }
\label{datasets}
\subsubsection{Dataset Preparation and Evaluation Metrics }
Our main experiments are conducted on the widely used Market-1501 \cite{zheng2015scalable} and DukeMTMC-reID \cite{ristani2016performance} datasets. Market-1501 dataset contains 32,668 images, belonging to 1501 identities. These images are taken from 6 cameras in front of the supermarket in the Tsinghua campus. Each identity is captured by multiple cameras. The standard evaluation protocol uses 12,936 bounding boxes of 751 identities for training and 19,281 bounding boxes of 750 identities for testing. DukeMTMC-reID dataset contains 36,411 images, belonging to 1812
identities and captured from 8 cameras. It provides 16,522 images of 702 identities for training and 19,889 images of 1110 identities for testing.
\emph{To distinguish the standard setting with the following SCT setting, we denote the standard Market-1501 protocol and DukeMTMC-reID protocol as Market-STD, Duke-STD, respectively.}

We construct SCT benchmarks following \cite{zhang2020single}. Specifically, for the training set, we randomly choose one camera for each person, and only take their images under the selected camera as training images. In this case, each person can only appear in one camera, so that there are no CCSP data in the training set. For the testing set, we keep the same as the standard testing set. We denote the SCT dataset built from Market-1501 and DukeMTMC-reID as Market-SCT and Duke-SCT, respectively.

Note that the number of training images is significantly reduced in the SCT training set due to the abandonment of CCSP data. Therefore, for fairly comparing, we propose a control group setting.
Take Market1501 as an example; we randomly selected the same number of images as Market-SCT from the standard Market-1501 training set, denoted as Market-CG. The same goes for building Duke-CG. The constructions of the training set of the SCT setting and control group setting are shown in Fig. \ref{sct_illu}.
The detailed statistics of the three settings are displayed in Tab. \ref{dataset}. 
Despite the above benchmarks, we also evaluate our method on the MSMT-SCT dataset \cite{ge2021cross}, which is derived from the MSMT-17 \cite{wang2018transferable} following \cite{ge2021cross}. The details are also reported in Tab. \ref{dataset}.

In this paper, we use two evaluation metrics, named Cumulated Matching Characteristics (CMC) and mean average precision (mAP). The CMC curve generally considers re-ID as a ranking problem and focuses on precision, while the mAP considers both precision and recall. We experiment with the above datasets and report mAP and the cumulated matching accuracy at Rank-1, Rank-5, and Rank-10.

\subsubsection{Model} 
Our method does not restrain the architecture of the deep network and can be applied to any model. We here take the ResNet-50 model pre-trained on ImageNet \cite{deng2009imagenet} as the backbone network to implement CIMN, considering its wide adoption in academia and industry. The four res-convolution blocks of the Resnet-50 model, stage1,stage2, stage3 and stage4, with a global maximize pooling layer, are adopted as the backbone. A Batch Normalization and a fully connected classifier with desired dimensions (corresponding to the number of identities of a person) are added in turn.
As shown in Fig. \ref{modelst}, the triplet loss within the cross-camera simulation and the meta-triplet loss are calculated by the features after the global maximize pooling layer.
The meta classification loss is calculated with the features after the classifier. For the meta camera alignment loss, we align the distribution of feature after stage2 between the meta-train set and the meta-test set.

\subsubsection{Implementation Details}

In this paper, all experiments are conducted with PyTorch.
The image size is $384\times128$ and padded with 10. For each minibatch, we randomly sample $P$ person from $2$ different cameras, respectively, and then randomly sample $K$ images of each person. Here
we set $P=8$ and $K=2$ so that the mini-batch size is $32$.
Some commonly used tricks, including random cropping, random flipping, random rotation, and random color jitter, are used. 
The trade-off coefficient of simulation loss $\lambda$ is set to 0.6. The weights of meta triplet loss, meta classification loss and meta alignment loss, $\gamma_1$, $\gamma_2$ and $\gamma_3$, are set to $1.0$, $1.0$ and $0.02$, respectively. We train the model for 240 epochs.
The learning rate $\eta$ is adjusted based on the epoch $t$ as follows. 
\begin{equation}
\label{eqlr}
\eta(t)=\left\{
\begin{aligned}
&3.5\times 10^{-4}\times \frac{t}{30}, &t\leq30\\
&3.5\times 10^{-4}, &5 \leq t\leq 120\\
&3.5\times 10^{-5}, &20 \leq t\leq180\\
&3.5\times 10^{-6}, &30 \leq t\leq240\\
\end{aligned}
\right.
\end{equation}

In addition, following two previous SCT re-ID works \cite{ge2021cross,wu2022camera}, we also implement our method on the global-local baseline, which additionally adds a local branch on the ResNet-50 (global branch) backbone. The implementation details are as follows. 1) In the global branch, the locations to use the meta triplet loss $L_{mtri}$, the meta classification loss $L_{mcl}$ and the meta camera alignment loss $L_{mca}$ are the same as our original one (see Fig. \ref{modelst}). 2)  $L_{mtri}$ and $L_{mca}$ are also adopted in the output of the local branch. 3) Other settings are the same as \cite{ge2021cross}. We denote this implementation as Ours(GL).

\subsection{Evaluation of CIMN}
This section evaluates the performance of our CIMN on Market-1501 and DukeMTMC-reID.
For comparison, we also train the baseline model with two main approaches in conventional supervised re-ID, i.e., trained with triplet loss and classification loss. We denote them as Bas-T and Bas-C, respectively.

\subsubsection{The Performance on Market-1501}
We first conduct an experiment on Market-1501. We trained the model on the training set of  Market-STD, Market-SCT and Market-CG respectively, and tested them on the standard testing set.
Tab. \ref{mar} reports the results. We have the following observations/conclusions:

(1) We can see the conventional approaches show a good performance on the Market-STD. The best performance is achieved by Bas-C, which hits 89.4\%, 96.7\%, 97.9\%, and 77.3\%  in Rank-1, Rank-5, Rank-10 and mAP, respectively.
However, if we train BAS-C on Market-SCT, it drop to 36.1\%, 53.5\%, 62.4\% and 13.2\%, with 53.3\%, 43.2\%, 35.5\% and 64.1\% performance degradations compared to Market--STD.
Similarly, the performance of Bas-T has also been suppressed by 50.4\%, 41.9\%, 35.7\% and 60.0\%.
In contrast, CIMN still remain 73.3\%/85.7\%/91.1\%/47.2\% in Rank-1/Rank-5/Rank-10/mAP accuracy, which significantly outperform Bas-C by 37.2\%/32.2\%/28.7\%/34.1\%. The results show a huge advantage of our CIMN in SCT re-ID.

(2) Since the number of training images is significantly reduced in the SCT training set, we evaluate these methods on Market-CG, where there is the same number of images as Market-SCT but contains many CCSP data. In this case, the Bas-C perform 73.9\%, 87.8\%, 93.4\%, and 47.9\%  accuracy in Rank-1, Rank-5, Rank-10 and mAP, respectively. And the performance of BAS-T is similar to it, both of which are  better than their results on Market-SCT. This phenomenon proves that the CCSP data is very  important for Bas-T and Bas-C that adopt conventional re-ID methods. 
On the other hand, the performance of CIMN on Market-SCT and Market-CG is very similar, and is comparable with that of conventional Bas-T/Bas-C on Market-CG. There are only 0.8\% and 0.9\% gaps in Rank-1 and mAP accuracy. \emph{This phenomenon demonstrates that by learning camera-invariant features with CIMN, the non-CCSP data and CCSP data in the training set can have a similar effect; thus, we can train the model without the expensive CCSP data, This is extremely valuable for the real-world application of re-ID.}

(3)We can see that CIMN also achieves better performance on Market-CG. It outperforms the second-placed Bas-C in Rank-1 and mAP by 0.2\% and 0.3\%. The results indicate that CIMN is also advantageous when training with a small number of images.

\begin{table}
\caption{ Performance on Market-STD, Market-SCT and Market-CG datasets. Rank-1 accuracy, Rank-5 accuracy, Rank-10 accuracy, and mAP (\%) are reported.}
\label{mar}
\centering
\begin{tabular}{lccccc}
\toprule

Dataset&Method & Rank-1 & Rank-5 & Rank-10 & mAP\\

\midrule
\multirow{3}{*}{Market-STD}&Bas-T&88.9&96.1& 97.4 &74.3 \\
&Bas-C&89.4&96.7& 97.9&77.3 \\
&CIMN&90.1&96.2& 97.5 &74.6 \\
\midrule
\multirow{3}{*}{Market-SCT}&Bas-T&38.5&54.2& 61.7 &14.3\\
&Bas-C&36.1&53.5& 62.4 &13.2 \\
&CIMN&73.3&85.7&91.1 &47.3 \\

\midrule
\multirow{3}{*}{Market-CG}&Bas-T&74.2&86.4& 92.1 &47.8 \\
&Bas-C&73.9&87.8& 93.4 &47.9 \\
&CIMN&74.1&87.9& 93.4 &48.2 \\
\bottomrule
\end{tabular}
\end{table}

\begin{table}
\caption{ Performance on Duke-STD, Duke-SCT and Duke-CG datasets. Rank-1 accuracy, Rank-5 accuracy, Rank-10 accuracy, and mAP (\%) are reported.}
\label{duke}
\centering
\begin{tabular}{lccccc}
\toprule

Dataset&Method & Rank-1 & Rank-5 & Rank-10 & mAP\\

\midrule
\multirow{3}{*}{Duke-STD}&Bas-T&81.5&91.6& 94.6 &65.6 \\
&Bas-C&83.2&91.3& 93.7&68.2 \\
&CIMN&81.7&91.3& 92.5 &65.9 \\

\midrule
\multirow{3}{*}{Duke-SCT}&Bas-T&20.7&34.1& 39.7 &11.6 \\
&Bas-C&24.7&31.4& 38.9 &12.1\\
&CIMN&68.4&79.2&84.1 &47.7 \\

\midrule
\multirow{3}{*}{Duke-CG}&Bas-T&68.5&83.1&87.5 &48.2 \\
&Bas-C&68.1&82.4&88.6 &48.6 \\
&CIMN&68.9&83.8& 88.7 &48.9 \\
\bottomrule
\end{tabular}
\vspace{-4mm}
\end{table}

\begin{table}
\setlength{\abovecaptionskip}{0cm}
\setlength{\belowcaptionskip}{0.5cm}
\caption{ Performance on global-local (GL) baseline. Rank-1 accuracy and mAP (\%) are reported.}
\label{GL}
\centering

\begin{tabular}{lcccc}
\toprule
\multirow{2}{*}{Method}&\multicolumn{2}{c}{ Market-SCT} &\multicolumn{2}{c}{ Duke-SCT } \\ 

\cmidrule(lr){2-3}\cmidrule(lr){4-5}
& Rank-1 & mAP & Rank-1& mAP\\

\midrule

Baseline(GL)\cite{ge2021cross} &  75.7& 51.5 &70.1 & 53.1\\

Ours(GL) & 85.6 &69.1 &81.0&64.5 \\

\bottomrule
\end{tabular}
\vspace{-4mm}
\end{table}






\begin{figure}

\centering
\hspace{-2mm}\subfigure[]{
\begin{minipage}[t]{0.45\linewidth}
\centering
\includegraphics[width=1.0\linewidth]{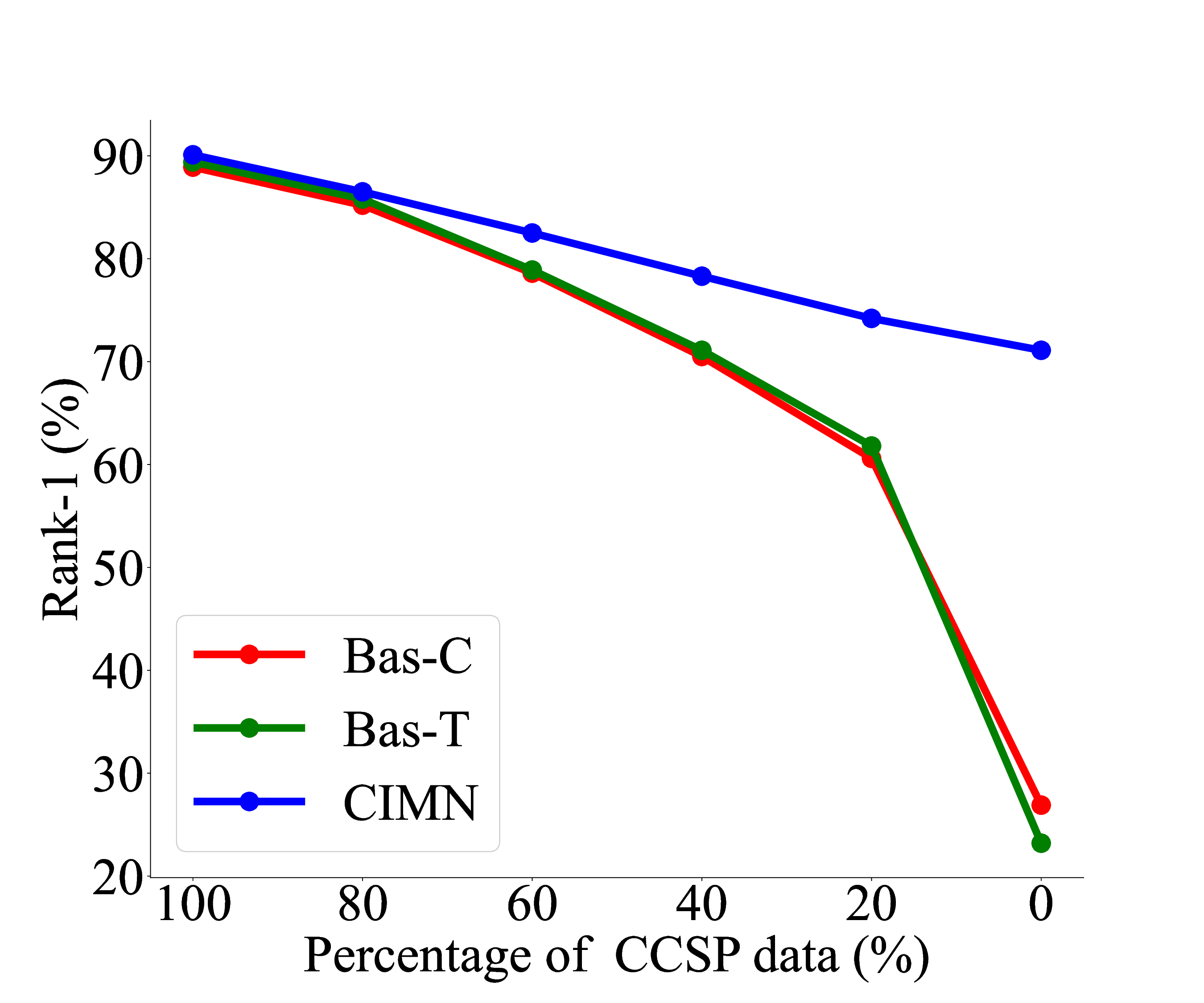}
\label{lamdasct}
\end{minipage}%
}%
\subfigure[]{
\begin{minipage}[t]{0.45\linewidth}
\centering
\includegraphics[width=1.0\linewidth]{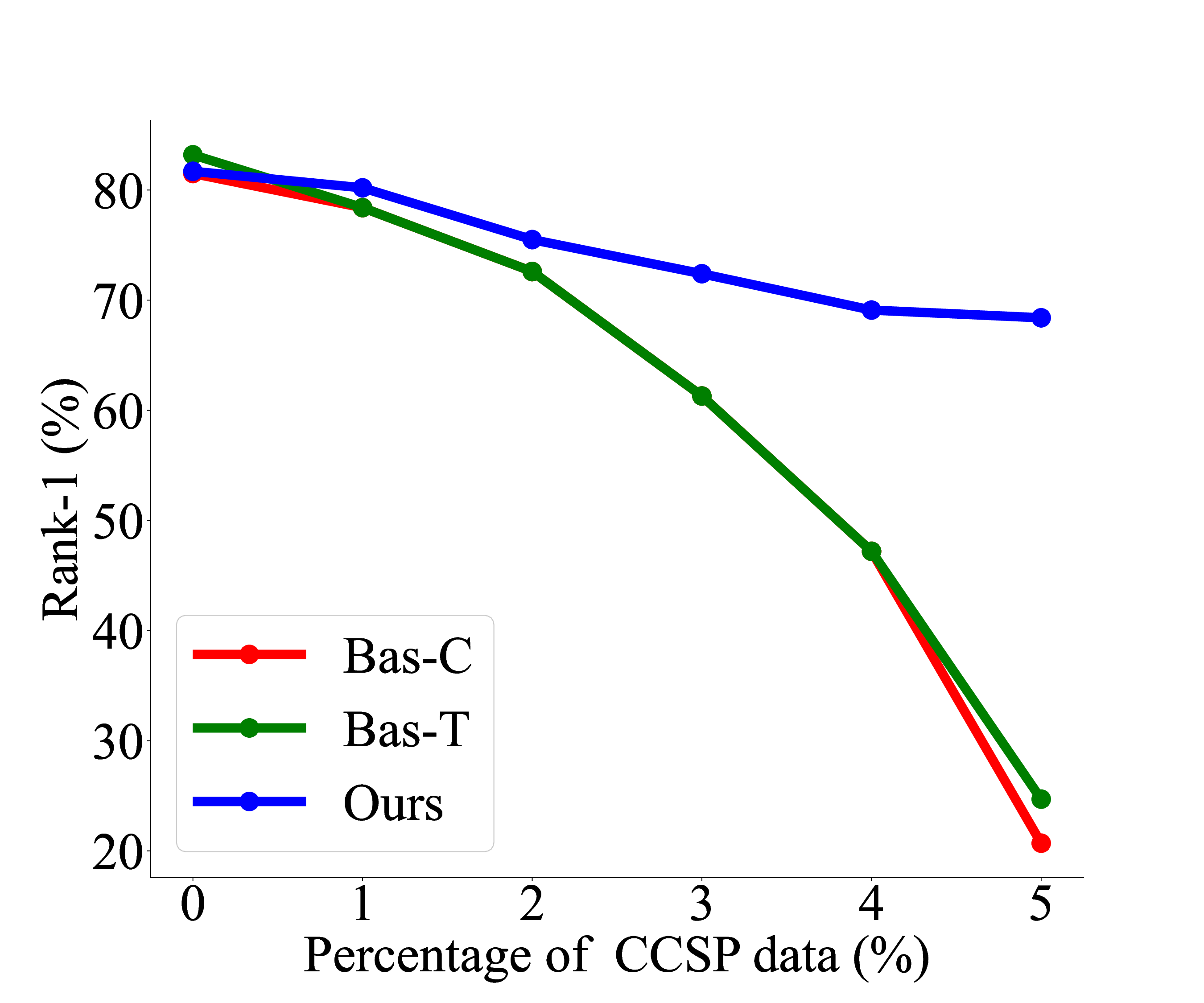}
\label{lamdasct}
\end{minipage}%
}%
\vspace{-2mm}
\caption{The performance tendency curves with changing CCSP data on Market-1501 dataset (Fig. a) and DukeMTMC-reID dataset (Fig. b).}
\label{sctcv}
\end{figure}

\subsubsection{Performance on DukeMTMC-reID}
We then evaluate our method on DukeMTMC-reID. Following the experiments on Market-1501, we conduct experiments on standard Duke-STD, Duket-SCT and Duke-CG.
From Tab. \ref{duke}, we can see our method is still effective.
While conventional methods (Bas-C, Bas-T) show poor performance on Duket-SCT due to the lack of CCSP data, CIMN still has 68.4\%, 79.2\%, 84.1\%, and 47.7\% accuracies in Rank-1, Rank-5, Rank-10 and mAP.
This outperform the two conventional methods in a large scale. For example, it surpasses Bas-C by 43.7\% in Rank-1 accuracy, 47.8\% in Rank-5 accuracy, 45.2\% in Rank-10 accuracy and 35.6\% in mAP. The improvements demonstrate the effectiveness of CIMN on SCT re-ID.
Besides, unlike conventional methods with large performance gaps on Duket-SCT and Duke-CG, CIMN achieves comparable performance on them. This phenomenon shows the essence of CIMN that reduces the reliances of the model on CCSP data.

\subsubsection{Stability}
Although the amount of CCSP data is small, it is inevitable that some people appear in not only one camera in the real-world. To evaluate the robustness of our CIMN, we conduct experiments to show how the accuracy changes with respect to the number of CCSP data. Fig. \ref{sctcv} shows the Rank-1 tendencies on Market-1501 and DukeMTMC-reID. We can see that CIMN is quite robust and stable with the changes of CCSP data; thus, it is more suitable for practical application.
\subsubsection{Extension to different structure}
Our method  guides the feature representation to be camera-invariant by enforcing the representation learned from a particular camera to perform well on other cameras, which is model-agnostic and feasible to be used on different structures. Here, we conduct experiments to evaluate CIMN on different models. Specifically, we implement CIMN on the global-local baseline used in recent SCT re-ID works (\cite{ge2021cross,wu2022camera}), which additionally  adds a local branch on the ResNet-50 (global branch) backbone. The implementation details are as follows. 1) In the global branch, the locations to use the meta triplet loss $L_{mtri}$, the meta classification loss $L_{mcl}$ and the meta camera alignment loss $L_{mca}$ are the same as our original one (see Fig. \ref{modelst}). 2)  $L_{mtri}$ and $L_{mca}$ are also adopted in the output of the local-branch. 3) Other settings are the same as \cite{ge2021cross}. The results are shown in Tab. \ref{GL}. It can be seen that our method is also effective in the global-local baseline, which improves the Rank-1/map of the baseline by 9.9\%/17.6\% and 10.9\%/11.4\%.

\subsection{Comparison with the State-of-the-Art Methods}
\vspace{-2mm}
\label{comsota}
Here we compare our method with state-of-the-art methods in three SCT re-ID benchmarks: Market-SCT, Duke-SCT and MSMT-SCT.
We first compare our performance with conventional supervised methods, including Center Loss \cite{wen2016discriminative}, A-Softmax \cite{liu2017sphereface}, ArcFace\cite{deng2019arcface}, PCB\cite{sun2018beyond}, Suh's method \cite{suh2018part},  MGN \cite{wang2018learning}, CBN \cite{zhuang2020rethinking}, HOReID \cite{wang2020high}, ISP \cite{zhu2020identity}, MGN-ibn \cite{wang2018learning}, Bagtrick \cite{joulin2016bag} and AGW \cite{ye2021deep}.
The results are reported in Tab \ref{sctc}. We can see that our method significantly outperforms these methods, which outperform the second one 11.4\%/6.7\% (CBN \cite{zhuang2020rethinking} on Market-SCT), 5.7\%/5.6\% (CBN \cite{zhuang2020rethinking} on Duke-SCT) and 0.3\%/1.2\% (MGN-ibn \cite{wang2018learning} on MSMT-SCT) respectively. Meanwhile, CBN is a transductive module that requires collecting test data to update the model, while our approach can be directly applied.

We then compare our method with the  SCT re-ID works MCNL\cite{zhang2020single}, which is also implemented on ResNet-50 baseline.
From Tab \ref{sctc}, we can see that CIMN is superior to their MCNL, which surpasses MCNL by 7.1\%/6.7\% Rank-1/mAP on Market-SCT, 2.0\%/2.4\% on Duke-SCT and 1.5\%/2.6\% on MSMT-SCT.

Finally, we compare our method implemented on the global-local baseline, i.e. Ours(GL) in Tab \ref{sctc}, with  two  SCT re-ID works which are also implemented on the global-local baseline: CCFP\cite{ge2021cross} and CCSFG \cite{wu2022camera}.
It can see that CIMN is competitive with CCFP\cite{ge2021cross} and CCSFG \cite{wu2022camera} on Duke-SCT, and achieves the best results on Market-SCT and MSMT-SCT, which surpasses the second one (CCSFG) by 0.7\%/1.5\% and  0.6\%/0.7\% Rank-1/mAP on the two datasets, respectively.

\begin{table}
\setlength{\abovecaptionskip}{0cm}
\setlength{\belowcaptionskip}{0.5cm}
\caption{ Compared with state-of-the-art in SCT re-ID setting. Ours(GL) denotes implementing CIMN on the global-local baseline. Rank-1 accuracy and mAP (\%) are reported. The best results are bolded.}
\label{sctc}
\centering
\resizebox{\linewidth}{!}{
\begin{tabular}{lcccccc}
\toprule
\multirow{2}{*}{Method}&\multicolumn{2}{c}{ Market-SCT} &\multicolumn{2}{c}{ Duke-SCT }&\multicolumn{2}{c}{ MSMT-SCT } \\ 

\cmidrule(lr){2-3}\cmidrule(lr){4-5}\cmidrule(lr){6-7}
& Rank-1 & mAP & Rank-1& mAP& Rank-1& mAP\\

\midrule

Center Loss \cite{wen2016discriminative} &40.3& 18.5& 38.7 &23.2&-&-\\
A-Softmax \cite{liu2017sphereface}&41.9 &23.2&34.8 &22.9&-&- \\
ArcFace \cite{deng2019arcface}&39.4 &19.8& 35.8 &22.8&-&-\\
PCB \cite{sun2018beyond}&43.5 &23.5&32.7 &22.2&-&- \\
Suh’s method \cite{suh2018part}&48.0& 27.3&38.5& 25.4&-&- \\
MGN \cite{wang2018learning}&38.1&24.7&27.1&18.7&-&-\\
CBN\cite{zhang2020single}&61.9&40.6&62.7&42.1&-&-\\
HOReID\cite{wang2020high} &48.1&29.6&40.2&29.8&-&-\\
ISP\cite{zhu2020identity} &60.1&40.5&61.4&41.3&-&-\\
MGN-ibn \cite{wang2018learning} &45.6&26.6&46.7&32.6&27.8&11.7\\
Bagtrick \cite{joulin2016bag}&54.0&34.0&54.2&40.2&20.4&9.8\\
AGW \cite{ye2021deep} & 56.0 & 36.6 &56.5 & 43.9 &23.0  &11.1\\
\midrule
MCNL\cite{zhang2020single} &66.2&40.6&66.4&45.3&26.6&10.0\\
Ours & \textbf{73.3} &\textbf{47.3} &\textbf{68.4}&\textbf{47.7}&\textbf{28.1}&\textbf{12.9} \\
\midrule
CCFP\cite{ge2021cross} &  82.4& 63.9 &80.3 & 64.5& 50.1 & 22.2\\
CCSFG \cite{wu2022camera}& 84.9 &67.7&\textbf{81.1}&63.7& 54.6& 24.6\\
Ours(GL) & \textbf{85.6} &\textbf{69.1} &81.0&\textbf{64.5}&\textbf{55.2}&\textbf{25.0}  \\

\bottomrule
\end{tabular}}
\vspace{-4mm}
\end{table}

\begin{table}
\caption{ Performance of the proposed solutions with the changing CCSP data on Market-1501 dataset. Rank-1 accuracy, and mAP (\%) are reported.}
\label{sctm}
\centering
\resizebox{\linewidth}{!}{
\begin{tabular}{lcccccccccc}
\toprule
Percentage of &\multicolumn{2}{c}{ CCS}&\multicolumn{2}{c}{+$L_{mtri}$ }&\multicolumn{2}{c}{ +$L_{mtri}$+$L_{mcl}$}&\multicolumn{2}{c}{+$L_{mtri}$+$L_{mcl}$+$L_{mca}$ } \\ 
\cmidrule(lr){2-3}\cmidrule(lr){4-5}\cmidrule(lr){6-7}\cmidrule(lr){8-9}\cmidrule(lr){10-11}
CCSP data & Rank-1 & mAP & Rank-1 & mAP& Rank-1 & mAP & Rank-1 & mAP\\

\midrule

100\%(Market-STD)&85.5&71.1&89.8&72.4&90.1&74.4& 90.7 &74.6 \\
80\% &82.1&64.2&86.4&69.2 &86.2&67.3& 86.5 &69.6\\
60\% &78.6&58.4&82.7&63.1&82.8& 62.9 & 82.9 &63.5\\
40\% &72.1&51.8&77.9&57.1&76.1&54.0& 78.3 &57.6\\
20\% &70.8&48.1&73.8&51.6&74.1&50.2&74.2 &51.4\\
0\%(Market-SCT) &68.4&44.9&70.7&46.5&72.3&46.9& 73.3 &47.3\\
\midrule
Market-CG &69.5&46.0&72.1&47.5&73.3&47.6&74.1&48.2\\
\bottomrule
\end{tabular}}
\vspace{-4mm}
\end{table}

\begin{table}
\caption{ Performance of the proposed solutions with the changing CCSP data on DukeMTMC-reID dataset. Rank-1 accuracy, and mAP (\%) are reported. }
\label{sctd}
\centering
\resizebox{\linewidth}{!}{
\begin{tabular}{lcccccccc}
\toprule
Percentage of &\multicolumn{2}{c}{ CCS}&\multicolumn{2}{c}{+$L_{mtri}$ }&\multicolumn{2}{c}{ +$L_{mtri}$+$L_{mcl}$}&\multicolumn{2}{c}{+$L_{mtri}$+$L_{mcl}$+$L_{mca}$ } \\ 
\cmidrule(lr){2-3}\cmidrule(lr){4-5}\cmidrule(lr){6-7}\cmidrule(lr){8-9}
CCSP data & Rank-1 & mAP & Rank-1 & mAP& Rank-1 & mAP & Rank-1 & mAP\\

\midrule

100\%(Duke-STD) &77.7&61.7&80.5&63.9&81.1&64.8& 81.7 &65.9 \\
80\% &76.5&57.5&79.1&61.7& 79.7& 61.8& 80.2 &62.1\\
60\% &70.8&54.9&75.1&59.7& 75.4& 59.5& 75.5 &60.5 \\
40\% &67.1&51.4&71.5&55.9& 71.3& 56.1& 72.4 &56.3\\
20\% &64.6&46.8&68.2&51.3& 68.9& 51.4& 69.1 &51.9\\
0\%(Duke-SCT) &62.4&41.0&65.2&45.2& 67.7& 46.9& 68.4 &47.7\\
\midrule
Duke-CG&62.9&41.7&64.6&45.8&67.8&46.9&68.9&48.9\\
\bottomrule
\end{tabular}}
\vspace{-4mm}
\end{table}

\begin{table}
\caption{ The performance of meta camera alignment loss at different positions. We evaluate the model on Market-SCT. Rank-1 accuracy, Rank-5 accuracy, Rank-10 accuracy, and mAP (\%) are reported.}
\label{ab1}
\centering
\begin{tabular}{lccccc}
\toprule
& Rank-1& Rank-5& Rank-10 & mAP\\

\midrule
Stage2 &73.3&85.7& 91.1 &47.3 \\
Stage3 & 73.1&85.2& 91.0&46.9 \\
Stage4 & 70.2&83.4& 89.1&42.7 \\

\bottomrule
\end{tabular}
\vspace{-4mm}
\end{table}

\begin{table}
\caption{ The influence of the number of  cameras $r$ in the meta-train set and the meta-test set on Market-SCT. If $r \leq 3$, the cameras in the two sets is completely different. Otherwise, there will be $2*r-6$ overlapped cameras in the two sets.}
\label{m_c}
\centering
\begin{tabular}{lccccc}
\toprule
r& Rank-1& Rank-5& Rank-10 & mAP\\

\midrule
1 &73.3&85.7& 91.1 &47.3 \\
2 & 72.8&85.2& 89.8&46.7 \\
3 & 72.7&85.2& 89.9&46.4 \\
4 & 66.9&81.4& 84.9 &40.1 \\
5 & 63.9&80.6& 84.7&38.5 \\
6 & 63.1&78.8& 84.7&37.9 \\
\bottomrule
\end{tabular}
\vspace{-4mm}
\end{table}
\vspace{-5mm}
\subsection{Ablation Experiments}
\label{ablation}
We further conduct the ablation experiments to evaluate the impact of the components of the proposed method. 
We first implement Cross-Camera Simulation to the model. Then we  add  the meta triplet loss $L_{mtri}$, the meta classification loss $L_{mcl}$ and the meta camera alignment loss $L_{mca}$ gradually.
We train it on the standard dataset (i.e., 100\% CCSP data) and then reduced the CCSP data until the CCSP data volume was zero. The result on Market-1501 and DukeMTMC-reID are reported on Tab. \ref{sctm} and Tab. \ref{sctd}, respectively.

\subsubsection{Effectiveness of Cross-Camera Simulation}
\label{dcce}
Cross-Camera Simulation (CCS) simulates a cross-camera process via meta-learning.
As shown in Tab. \ref{sctm} and Tab. \ref{sctd}, with CCS, the model is more  robust and stable with the changes of CCSP data. If we throw away all the CCSP data, i.e., on Market-SCT and Duke-SCT, it remains 68.4\%/44.9\% and 62.4\%/41.0\% Rank-1/mAP accuracy. This result is close to its performance on Market-CG (69.5\%/46.0\%) and Duke-CG (62.9\%/41.7\%), indicating the dependence of the model on CCSP data is significantly reduced.

However, since CCS separates different cameras, this strategy can not fully mine the relations between the data from these cameras.
As a result, the performance of the model on Market-STD/Duke-STD and Market-CG/Duke-CG is worse than that of conventional approaches Bas-T and Bas-C (reported in Tab. \ref{mar} and Tab. \ref{duke}).

\subsubsection{Effectiveness of Meta Triplet Loss }

Meta triplet loss $L_{mtri}$ guides the model to leverage the potential negative relations between meta-train set and meta-test set.
With $L_{mtri}$, the performance of the model on Market-STD and Duke-STD improves 4.3\% and 2.8\% in Rank-1 accuracy, respectively.
Similarly, its Rank-1 accuracy on Market-CG and Duke-CG are improved by 2.6\% and 1.7\%. These results demonstrate the effectiveness of $L_{mtri}$.
Besides, we can see that on Market-SCT and Duke-SCT, $L_{mtri}$ also improves CCS by 2.3\% and 2.8\% in Rank-1 accuracy. 

\subsubsection{Effectiveness of Meta Classification Loss }
We then add meta classification loss $L_{mcl}$ to our model, with the same purposes as $L_{mtri}$.
From Tab. \ref{sctm} and Tab. \ref{sctd}, we can see that the $L_{mcl}$ is effective, which improves the Rank-1 accuracy by 0.3\%/0.6\% on Market-STD/Duke-STD and 1.2\%/3.2\% on Market-CG/Duke-CG.
On SCT setting, it also brings 1.6\% and 2.5\% improvements on Rank-1 accuracy on Market-SCT and Duke-SCT.

\subsubsection{Effectiveness of Meta Camera Alignment Loss }
Finally, we conduct meta camera alignment Loss $L_{mca}$ to our model, which actively assigns the distribution of meta-train set and meta-test set, which encourage the model to leverage the potential positive relations between the two sets.
We can see that $L_{mtri}$ improves the Rank-1/mAP accuracy by 0.6\%/0.2\% and 0.6\%/1.1\% on Market-STD and Duke-STD.
It is also beneficial to the SCT re-ID setting, which improves the Rank-1/mAP of the model by 0.8\%/0.6\% and 1.1\%/2.0\% on Market-SCT and Duke-SCT, respectively.

\subsubsection{Influence of the Number of Cameras in Meta-train and Meta-test Processes}
In our Meta-Batch Preparation, we collect samples from one camera to the meta-train set and that from another camera to the meta-test set. Here we explore the result if there are multiple cameras in two sets.
Specifically, in each meta-batch, we randomly select $r$ cameras for the meta-train set and the meta-test set, respectively. 
If $r\leq N/2$, where $N$ is the number of the cameras in the dataset, the cameras in the meta-train set and the meta-test set are non-overlapped. Otherwise, there will be $2*r-N$ overlapped cameras  in the two sets simultaneously.
Other parts are the same as our original implementation. Note that, although there are samples from multiple cameras in the meta-train/meta-test set, there is no overlapped person among these cameras, i.e., we still guarantee that each person in the training set only appears in one camera to satisfy the SCT setting.

The results are reported in  Tab. \ref{m_c}. We can see that if there are overlapped cameras in the two sets, the performance of the model will degrade significantly. If the camera composition is completely the same, i.e. $r=6$, the mAP drops from $46.1\%$ to $36.9\%$. We hypothesize the reason is that the meta-learning-based cross-camera simulation process encourages the model to learn the knowledge that is suitable for the meta-train set and meta-test set. In other words, this implicitly guides the model to learn to bridge the gap between the two sets. Therefore, the larger the gap between the camera composition of the meta-training set and the camera composition of the meta-test set, the more our method can promote the model to learn camera-invariant feature representations to bridge the camera gap. In addition, it can be seen that $r=1$ is better than other cases $r=2,3$, but very slightly. We argue this is because when $r=1$, our $L_{mca}$ can better capture the feature distribution of the data from each camera and align them.

\begin{figure}[tbp]
\centering

\subfigure[]{
\begin{minipage}[t]{0.45\linewidth}
\centering
\includegraphics[width=1.0\linewidth]{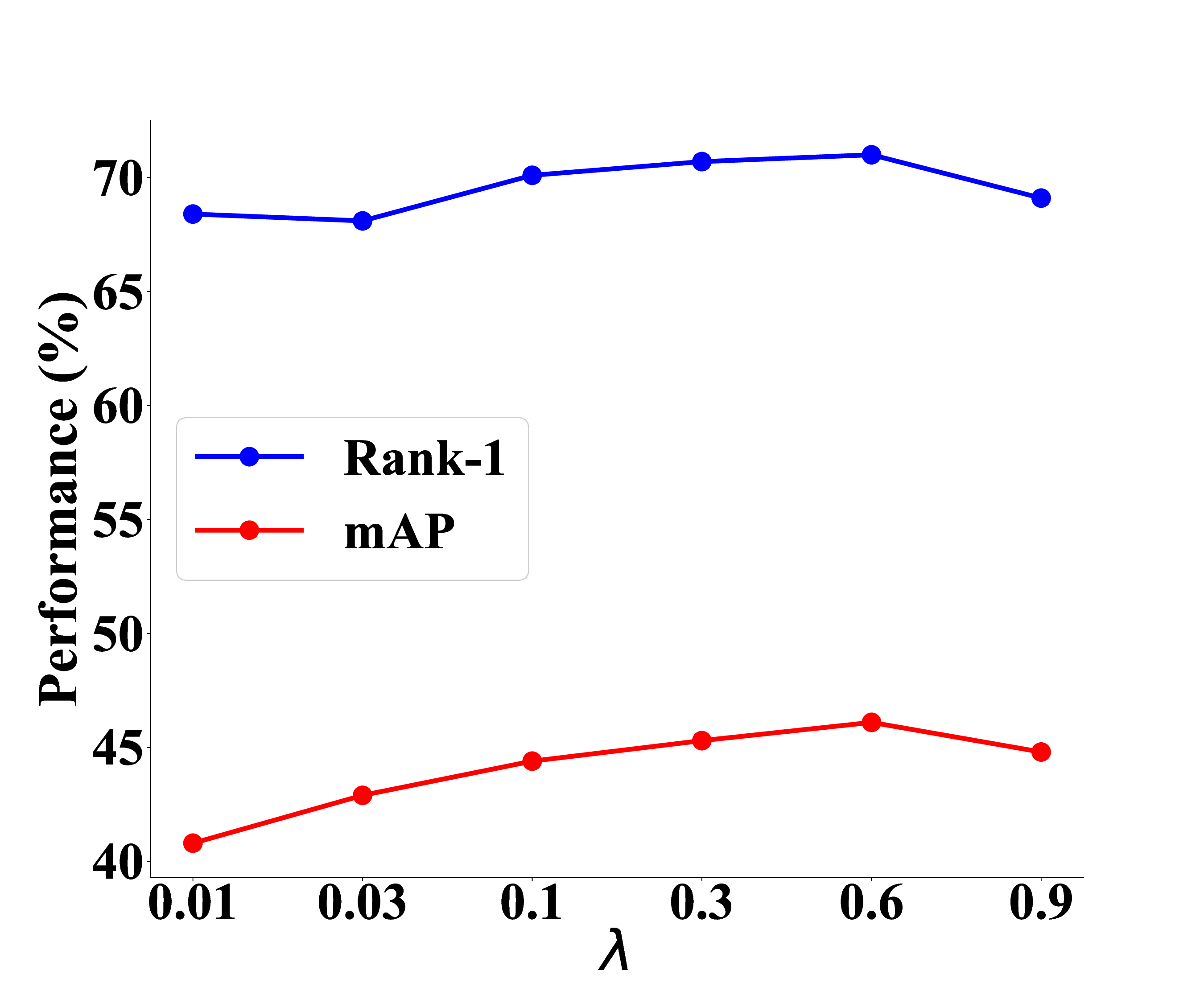}\label{paraa}
\end{minipage}%
}
\subfigure[]{
\begin{minipage}[t]{0.45\linewidth}
\centering
\includegraphics[width=1.0\linewidth]{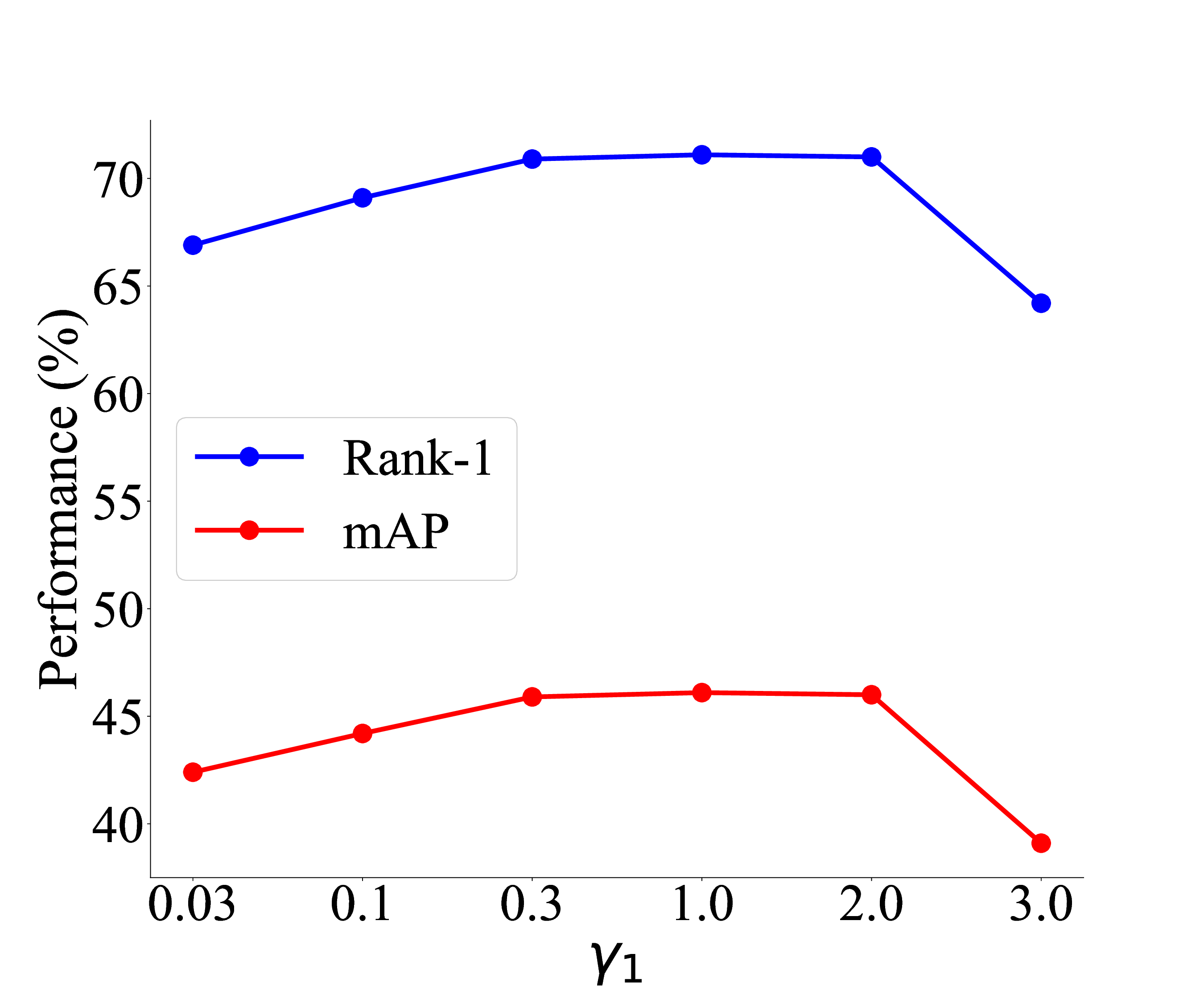}\label{parab}
\end{minipage}%
}\vspace{-2mm}
\subfigure[]{
\begin{minipage}[t]{0.45\linewidth}
\centering
\includegraphics[width=1.0\linewidth]{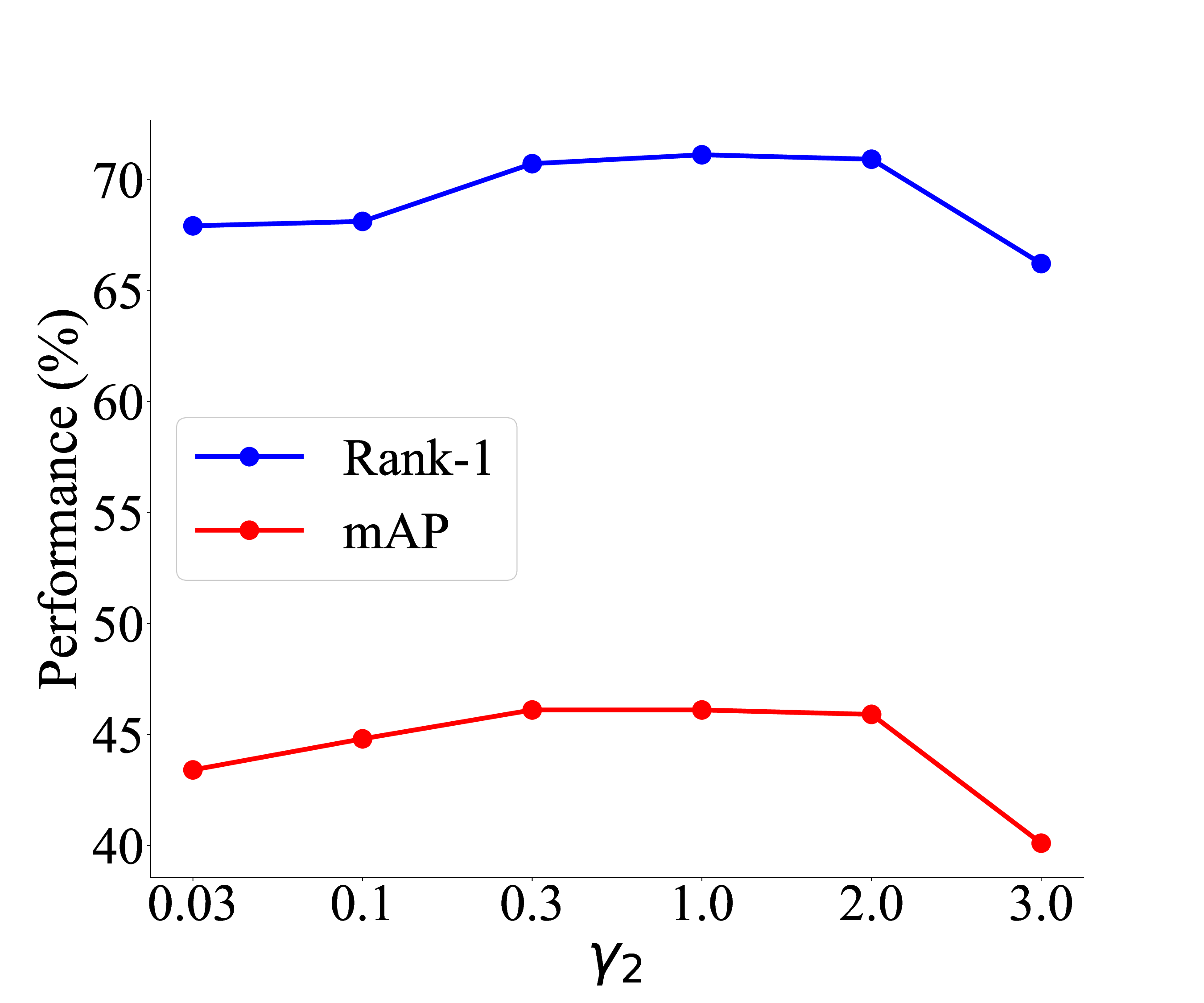}\label{parac}
\end{minipage}%
}
\subfigure[]{
\begin{minipage}[t]{0.45\linewidth}
\centering
\includegraphics[width=1.0\linewidth]{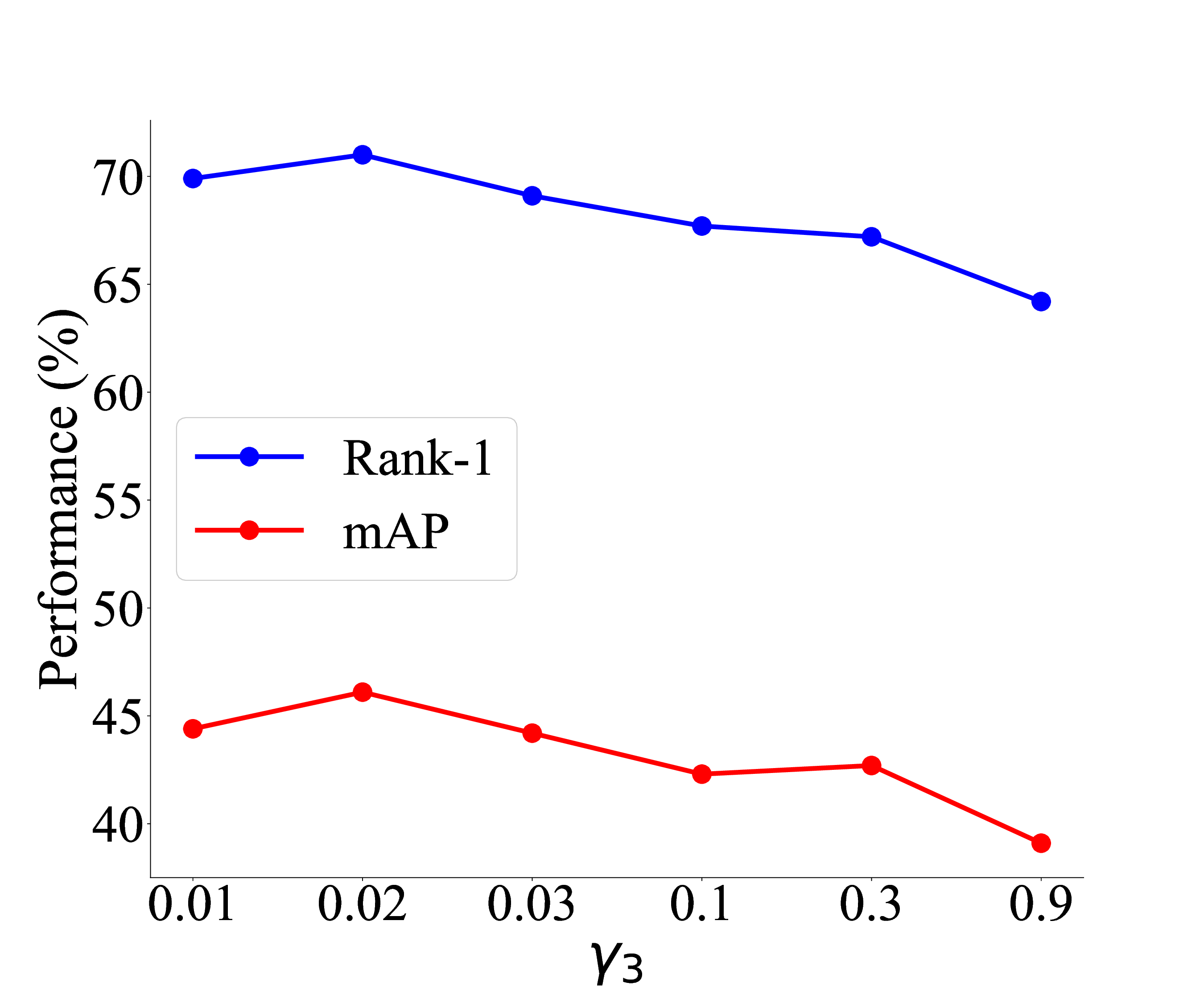}\label{parad}
\end{minipage}%

}
\centering
\vspace{-2mm}
\caption{The Rank-1/mAP Curves with different $\lambda$, $\gamma_1$, $\gamma_2$ and $\gamma_3$ on Market-SCT. }
\vspace{-2mm}
\end{figure}
\vspace{-2mm}




\vspace{-1mm}
\subsection{Design Choices}
\vspace{-2mm}
\label{desc}
Here we illustrate our design choices, including the selection of hyperparameters and the position to implement meta camera alignment loss. We evaluate our method on Market-SCT.
\subsubsection{Important Hyperparameters }
We examine the impact of trade-off coefficient $\lambda$ and the weight of meta triplet loss $\gamma_1$, meta classification loss $\gamma_2$ and meta camera alignment loss $\gamma_3$, respectively. For each hyperparameter, we fix other hyperparameters and then adjust this hyperparameter in a wide range. 
The results are shown in Fig. \ref{paraa}, it can be seen that the best performance is achieved when $\lambda=0.6$.

From Fig. \ref{parab} and Fig. \ref{parac}, we note that the choice of $\gamma_1$ and $\gamma_2$ has little effect on the model within a certain range (almost in [0.3, 2.0]).
Therefore, we set $\gamma_1$ and $\gamma_2$ to 1.0 for convenience.
The performance curve of  $\gamma_3$ is shown in Fig. \ref{parad}. It can be seen that the best performance is achieved when $\gamma_3=0.02$. 

\subsubsection{Where to Apply Meta Camera Alignment Loss }
Meta camera alignment loss $L_{mca}$ aims to align the feature distributions between different cameras.
Here we discuss where $L_{mca}$ loss is applied.
For simplicity, we only consider the following four simple cases: aligning the distribution of the feature after stage2, stage3, and stage4. We find that the impose the loss on the shallow layer is more effective than the deep layer. As shown in Tab \ref{ab1}, the best performance is achieved on stage2. We believe that the reason is that: 1) $L_{mca}$ is not related to the specific identity of the data, so it does not need to be added to the deep layer of the network model where the output features represent the identity information of data. 2) By aligning the feature distributions of data from different cameras, $L_{mca}$ can help to filter out camera-related features. These camera-related features are more likely to be low-level semantics so that can be better handled in shallow layers.

\begin{table}
\caption{ Compared with state-of-the-art in single domain generalization re-ID setting.
Label M, D denote Market-1501 dataset and DukeMTMC-reID dataset, respectively. M$\rightarrow$D means training the model on Market-1501 and testing it on DukeMTMC-reID.
Rank-1 accuracy and mAP (\%) are reported. The best results are bolded.}
\label{g1}
\centering
\begin{tabular}{lcccc}
\toprule
\multirow{3}{*}{Method}&\multicolumn{2}{c}{ M$\rightarrow$D} &\multicolumn{2}{c}{ D$\rightarrow$M } \\ 

\cmidrule(lr){2-3}\cmidrule(lr){4-5}
& Rank1 & mAP & Rank1& mAP\\

\midrule

StrongBaseline \cite{luo2019strong}&41.4&25.7&54.3&25.5\\
OSNet-IBN \cite{zhou2019learning}&48.5&26.7&57.7&26.1\\
OSNet-AIN \cite{zhou2019learning}&52.4&30.5&61.0&30.6\\
PAP \cite{huang2018enhancing}&46.4&27.9 &59.5&30.6 \\
SNR \cite{jin2020style}&55.1 &33.6&66.7&33.9 \\
DIR-ReID \cite{zhang2021learning}&54.5 &33.0&\textbf{68.2}&\textbf{35.2} \\
\midrule
CIMN & \textbf{58.3} &\textbf{36.7} & 65.9& 32.9\\

\bottomrule
\end{tabular}
\vspace{-4mm}
\end{table}
\vspace{-5mm}
\subsection{Generalization Ability}
In addition to our motivation to address SCT re-ID, CIMN is robust to camera changes since the learned representations are camera-invariant. As a result, the model is more generalizable to new re-ID datasets captured by new cameras. To verify this point, we evaluate our CIMN on single domain generalization re-ID setting, where the model is trained on a source dataset and tested on unseen target datasets.

Tab. \ref{g1} shows the comparison between our method and the state-of-the-arts methods in this field, including StrongBaseline \cite{luo2019strong}, OSNet-IBN \cite{zhou2019learning}, OSNet-AIN \cite{zhou2019learning}, PAP \cite{huang2018enhancing}, SNR \cite{jin2020style} and DIR-ReID\cite{zhang2021learning}.
As illustrated, our method is comparable with the state-of-the-art methods. Especially, we achieve the best performance on M$\rightarrow $D, which outperform the second one \cite{jin2020style} by 3.2\% and 0.6\% in Rank-1 accuracy and mAP, respectively.

We also compare our method in  Mutli-source generalization setting \cite{song2019generalizable,chen2020dual,jin2020style}. In this setting, the model trained on five source datasets: CUHK02 \cite{li2013locally}, CUHK03 \cite{li2014deepreid}, Market-1501, DukeMTMC-reID, and CUHK-SYSU \cite{xiao2016end} datasets, with a total of 121,765 images of 18,530 identities. And then the model is directly evaluated on four testing datasets: VIPeR \cite{gray2008viewpoint}, PRID \cite{hirzer2011person}, GRID \cite{loy2010time}, and i-LIDS \cite{2009Associating}.
However, there are no available camera annotations in the CUHK-SYSU dataset. Therefore, we only used 6596 identities from the four remaining datasets, i.e., CUHK02, CUHK03, Market-1501, and DukeMTMC-reID, as the training set.
Our method achieves 50.4\%/52.6\%/48.0\%/76.1\% mAP on the VIPeR/PRID/GRID/i-iLDS, which outperforms Bas-T by 6.0\%/16.9\%/12.2\%/9.6\% and   Bas-C 6.2\%/15.1\%/12.4\%/4.9\%, respectively. The above results prove the effectiveness of our method in improving the model's generalization ability.



\vspace{-4mm}
\section{Conclusion}
\label{sec:6}
\vspace{-2mm}
In this paper, we propose a Camera-Invariant Meta-Learning Network (CIMN) for Single-camera-training person re-identification. CIMN performs a cross-camera simulation by split the training data into meta-train set and meta-test set based on camera IDs, enforcing the representations learned from one set to be robust on the other set. Then three losses, meta triplet loss, meta classification loss and meta alignment loss, are introduced to leverage the relations ignored by the cross-camera simulation.
With the interaction from the meta-train set and the meta-test set, the model is guided to learn camera-invariant and identity-discriminative representations.
We demonstrate that our method achieves comparable performance with and without CCSP data, and outperforms the state-of-the-art methods on two SCT re-ID benchmarks.
Finally, we show that our method is also effective in improving the model's generalization ability.

\vspace{-4mm}
\section*{Acknowledgment}
\vspace{-2mm}
This work is sponsored by the National Key Research and Development Program under Grant (2018YFB0505200) and National Natural Science Funding (No.62002026).

\vspace{-2mm}
\bibliographystyle{IEEEtran}%
\bibliography{ref}

%







\end{document}